\documentclass[11pt]{article}
\usepackage{graphicx}
\begin{document}
\nopagebreak[4]
\title{Electrocardiography Separation of Mother and Baby}
\author{Wei Wang}
\maketitle
\begin{abstract}
Extraction of Electrocardiography (ECG or EKG)  signals of mother and baby is a challenging task, because one single device is used and it receives a mixture of multiple heart beats. In this paper, we would like to design a filter to separate the signals from each other. 
\end{abstract}
\section{Introduction}
Electrocardiography (ECG or EKG) is a transthoracic interpretation of the electrical activity of the heart over time captured and externally recorded by skin electrodes \cite{ECG}. Intrinsically the electrods add some noises to the final data captured and also some external sources, usually from the human body, add some other noises. Dealing with these sorts of noises is one of the challenges of recording the correct ECG of a person heart beat.\\
As a regular task ECG is used to record heart beat of a person, but in some especial cases it is utilized to record the heart beat of two persons, mother and the baby in her abdomen. In these cases the ECG gives us a mixture of two heart beats and the noises explained before. The important task will be to distinguish the two different heart beats. To help this issue be solved, several electrods are attached to the mother's body. Some of them are closer to her chest and some are closer to her abdomen. Since different electrodes capture different data, obviously each captures the closest source to itself as the strongest signal, utilizing all of them is a good solution for signal separation. \\
Filter in signal processing is commonly used to refer to any device or system that takes a mixture of frequency components from its input and process them according to some rules to generate a corresponding set of frequencies at its output \cite{Farhang}. Considering the situation we explained in the last paragraph, we aim to extract the child signal using 8 electrodes attached to mother's body, three on her chest and five on her abdomen. Using filters is one of the solutions to this problem. Obviously, the human's heart beat is not constant all the time. As the time goes on the heart beat changes for different reasons. That means for filtering it, an online and adaptive filter is required to follow the heart beat in time. \\
In this report we are to explain our approaches to extract child's heart beat from these eight electrodes. We tried two approaches that one of them was successful and gives us quite good results, while the implementation of the second one was failed because of some problems that will be explained later. The remainder of the report is organized as follows. Section two describes our approach in signal extraction. Section three is to present the results of the approach with different parameter settings. Section four gives us some idea about a filter. And finally the last section concludes everything in the paper. \\
\section{Our Approach}
\begin{figure}[t]
\centering
\includegraphics[scale=0.3]{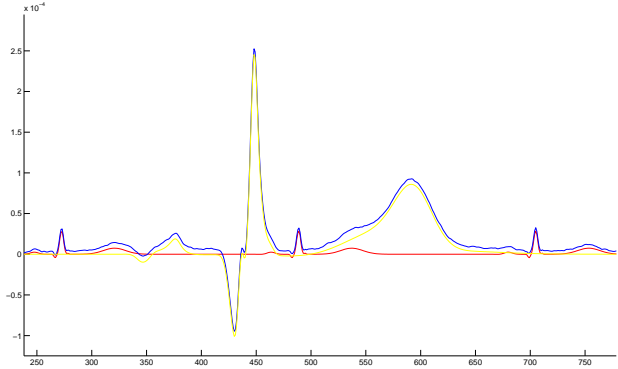}
\caption{Red is child signal, Yellow mother signal and Blue is the mixture of signals captured from the first electrode placed on mother's chest}
\label{elem}
\end{figure}

\begin{figure}[t]
\centering
\includegraphics[scale=0.4]{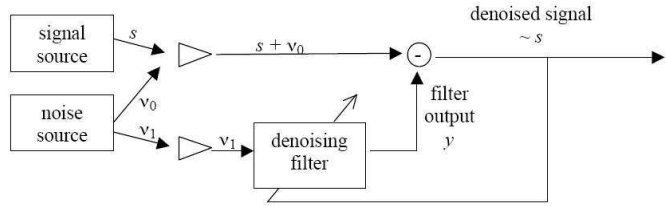}
\caption{Denoising filter, picture from \cite{Notes}}
\label{denoise}
\end{figure}

In Figure \ref{elem} we show three signals in the same diagram. Having the blue signal, we would like to extract the red signal. That means we can consider the rest as noise, both mother signal and real noises. In general a denoising filter works as Figure \ref{denoise}. The error that is going to be minimized in this type of filter is $error = s+\nu_0 - y$ \cite{Notes}.\\

Our approach is using Least Mean Square aalgorithm in filtering, to denoise the mixed signal and extract the child signal \cite{shams1, shams2}. Here we explain how we implemented the LMS algorithm. We defined L as the delay time. So we consider the last L input data in order to measure the current noise signal. We assigned weights,W , to all these input data and  made the output as the weighted some of these L elements.We call it the total noise at time t. By subtracting this noise from the mixed signal we end up with the child signal. The following equations were used in our model implementation where CS is the chest signal and AS is the abdomen signal.  \\
\[noise(t) = CS(t-L+1 : t) \cdot W \]
\[child(t) = AS(t) - y(t)\]
Now we update the weight array using the following equation where lr is the learning rate set by default to 600\\
\[W = W + 2lr\cdot child(t)\cdot CS(t-L+1:t)^T\]
We can apply this method on one chest signal and one abdomen signal to extract the child signal. But it is not so accurate and differences between the original signal and the measured one is very much. So we refined this approach by some simple steps. We apply it to two signals from chest and two from abdomen pairwise. Then we apply another filter on the output of these signals and try to find the average of them by our filter. So this time the output is not the differences between signals but the average that last time we mentioned as noise. Using this method we got to better results that is shown in the next section.\\
The other refinement that we did to this approach was to change the learning rate in time. So we used an approximation of auto-correlation matrix of the input signal. According to this matrix we can find a good approximation of learning rate. The equations of choosing lr is as follows where R is the correlation matrix, x is the input signal and M is the constant for misadjustment. It is notable that in this case we need the last J+L input data to measure the matrix. We call J as window size later.\\
\[D(t) = \left[ \begin{array}{cccc}
x(t)&x(t-1)& ... &x(t-L+1)\\
x(t-1)&x(t-2)& ... &x(t-L)\\
...  \\
x(t-J)&x(t-J-1)& ... &x(t-J-L+1)\\
\end{array}
\right]\]
\[R = D^T\cdot D\]
\[lr = \frac{1}{3trace(R)}\]
or
\[lr = \frac{M}{trace(R)}\]
If we choose the first learning rate, it converges very fast and usually it has lots of overshootings. The second one converges slower if $M<1/3$. So we can use the first learning rate at the begining of our run and the second one later. Within this criteria we use the benefits of both of these approaches. We tested our algorithm with different parameters that are shown in the next section.

\section{Experiment and Results}
\begin{figure}[t]
\centering
\includegraphics[scale=0.3]{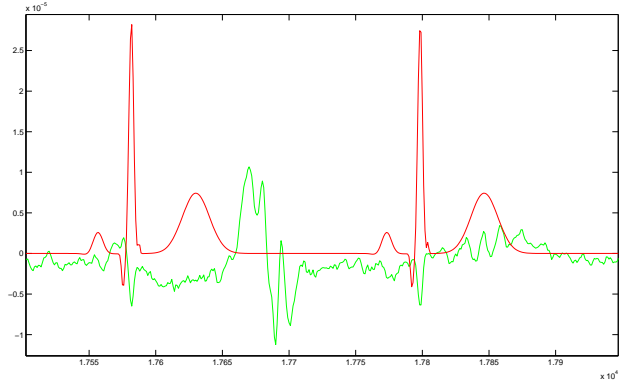}
\caption{Green is the computed signal and Red is the real child signal}
\label{result}
\end{figure}

We tested our approach by a sequence of mixed signal with the length of 90,000. This sequence is used to store the data for 3 minutes that gives us the frequency of 500Hz. We applied this sequence to our filter function. We tested the results with the real child signal that is given by Professor Jaeger, and is available on the course home page. Having the result and the real target signal ,the equation for computing error is as follow that considers the last 75,000 points.\\
\[\mbox{diff} =  \mbox{result} - \mbox{target} \]
\[\mbox{diff} = {X_1,X_2,...,X_n}\]
\[\mbox{error} = \sqrt{\frac{\frac{\Sigma X_i}{n}}{var(\mbox{target})}}\]
The first test was with only one filter that was provided in the sample code lr = 600.  The error was 3.7117. The rest of experiments were done using three filters but under different conditions. The first two filters were borrowed from the sample with this accuracy and the second filter is implemented to have changing learning rate. This means after a refinement by 3.7117 we try to apply the other filter to average these two signals. Table 1 present the results of applying several parameters on the second layer filter. We set the threshold of changing learning rate equation to 15,000. Within the first 15,000 iterations it does not use M constant at all.\\
\begin{table}[htp]
\centering
\begin{tabular}{cccc}
Delay Length(L) & Misadjustment(M) & Window Size (J) & Accuracy\\
\hline
1 & 10e-5 & 1 & 9.2580e16\\
1 & 10e-5 & 2 & 24.9195\\
1 & 10e-5 & 5 & 1.2631\\ 
1 & 10e-5 & 10 & 1.4084\\
1 & 10e-7 & 1 & 4.6396e5\\
1 & 10e-7 & 2 & 2.0496\\
1 & 10e-7 & 5 & 1.7636\\
1 & 10e-7 & 10 & 1.9344\\
2 & 10e-5 & 1 & 3.2803e24\\
2 & 10e-5 & 2 & 2.0559\\ 
2 & 10e-5 & 5 & 1.644\\
2 & 10e-5 & 10 & 1.5154\\
2 & 10e-7 & 1 & 9.6966e24\\
2 & 10e-7 & 2 & 1.3079\\
2 & 10e-7 & 5 & 1.6843\\
2 & 10e-7 & 10 & 2.0556\\
5 & 10e-5 & 1 & inf\\
5 & 10e-5 & 2 & 6.0334e106\\
5 & 10e-5 & 5 & 42.9274\\
5 & 10e-5 & 10 & 1.8316\\
5 & 10e-7 & 1 & inf\\
5 & 10e-7 & 2 & 2.0537e107\\
5 & 10e-7 & 5 & 64.3655\\
5 & 10e-7 & 10 & 2.0689\\
10 & 10e-5 & 1 & inf\\
10 & 10e-5 & 2 & inf\\
10 & 10e-5 & 5 & 2.1790e55\\
10 & 10e-5 & 10 & 6.9142e10\\
10 & 10e-7 & 1 & inf\\
10 & 10e-7 & 2 & inf\\
10 & 10e-7 & 5 & 4.6250e55\\
10 & 10e-7 & 10 & 7.6576e10\\
\end{tabular}
\caption{Different parameters and their results}
\label{tab:}
\end{table}
It can be seen in the table the best performance is achieved at L=1, M=10e5 and J=5. At that point the computed signal follows the child signal as in Figure \ref{result}. We see that increasing L usually decreases the performance but J acts vice versa. Although the best performance is achieved when M was 10e-5, the general performance of M = 10e-7 is better. \\

\section{An Unsuccessful Approach}
\begin{figure}[t]
\centering
\includegraphics[scale=0.25]{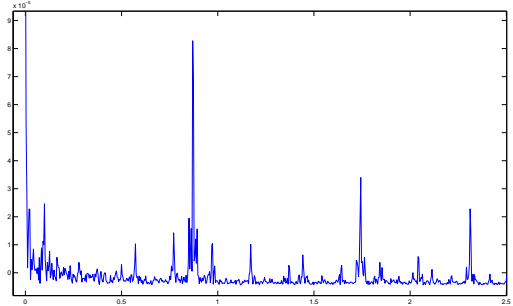}
\caption{The mixed signal in frequency domain. It is shown that the mother frequency is around 0.8 and for the child it is around 2.3}
\label{freq}
\end{figure}

The first idea that I had at the begining was using spectral analysis to separate two different signals from each other, but after spending some time on it, I could not finish. The idea was to use Discrete Fourier Transform(DFT) to separate two different signals. Since mother and child do not have the same beat rate, the frequency is different. DFT can transform signals from time domain to frequency domain. Using this transformation, I could extract the different frequencies and choose the child frequency that is greater than mother beat frequency. Then I would choose the smaller frequency and do the inverse fourier transform in order to go back to time domain. The other issue was the change of frequency in time. Thus, I decided to choose the last two seconds only and show the results with two seconds of delay. It is notable that less that one second was not possible, because the frequncy of mother goes less than one at some points in time. In Figure \ref{freq} the result of applying DFT to the mixed signal is shown. \\
The problem that I could not solve and stoped me in this approach was the noise. There are several noises in the frequency domain and it was not easy to find the one that is the child frequency. 
\section{Conclusion}
In this report we explained two filters for denoising an ECG signal in order to extract the child's heart beat from mother's heart beat and other noises. We showed the first and starightforward approach using adaptive filters. We explained our results with different settings on the same input data. There was another approach that was not successful but in theory it must work. It can be a future work to do this filtering by spectral analysis. However the delay is considerable. 

\end{document}